\documentclass[conference]{IEEEtran}
\IEEEoverridecommandlockouts
\usepackage{cite}
\usepackage{amsmath,amssymb,amsfonts}
\usepackage{algorithmic}
\usepackage{graphicx}
\usepackage{textcomp}
\usepackage{xcolor}
\usepackage{multirow}  
\usepackage{booktabs} 
\usepackage{graphicx} 
\def\BibTeX{{\rm B\kern-.05em{\sc i\kern-.025em b}\kern-.08em
    T\kern-.1667em\lower.7ex\hbox{E}\kern-.125emX}}
\begin{document}

\title{SABER: A Semantic-Aligned Brain Network Analysis Framework via Multi-scale Hypergraphs\\
\thanks{This work is supported in part by the Zhejiang Provincial ``Jianbing Lingyan+X'' Science and Technology Program (2025C01127). 
\textit{Corresponding author: Huihui Ye (yehuihui@hdu.edu.cn).}}
}

\author{
  \IEEEauthorblockN{
    Yidan Xu$^{1}$,
    Xiangmin Han$^{2}$,
    Rundong Xue$^{3}$,
    Huihui Ye$^{1,\dagger}$
  }
  \IEEEauthorblockA{
    $^{1}$ Hangzhou Dianzi University, China \\
    $^{2}$ Tsinghua University, China \\
    $^{3}$ Xi'an Jiaotong University, China \\
    Emails: YidanXu2024@163.com, simon.xmhan@gmail.com, xuerundong@stu.xjtu.edu.cn, yehuihui@hdu.edu.cn
  }
}

\maketitle

\begin{abstract}
Effective brain disease diagnosis requires the synergy of brain connectivity patterns and high-level semantic knowledge. Existing methods, however, largely treat semantics from large language models (LLMs) as auxiliary features or supervision, limiting their direct role in decision-making and constraining classification stability and robustness. To overcome this, we propose a semantic-aligned brain network framework that actively integrates LLM-derived semantics into the prediction process. Specifically, ROI-level semantics are first incorporated via global self-attention to enrich node representations and provide whole-brain context. Multi-scale hypergraphs are then constructed to explicitly model functional subnetworks and multi-ROI interactions, addressing the locality limitations of traditional GNNs and capturing high-order dependencies. Finally, a decision-level semantic alignment mechanism selectively injects patient-specific textual embeddings into graph representations, enabling semantics to directly guide predictions without perturbing the underlying network structure. Experiments on public brain network datasets ABIDE and ADHD-200 demonstrate state-of-the-art performance, enhanced stability, and improved interpretability, particularly in small-sample settings.
%Brain network-based disease diagnosis relies not only on neural connectivity patterns but also on high-level semantic knowledge. However, existing methods primarily treat the semantics provided by large language models (LLMs) as auxiliary features or supervisory signals, which limits their direct involvement in the decision-making process and constrains classification stability and robustness. To address this limitation, we propose a semantic-aligned brain network modeling framework that actively incorporates LLM-derived semantics into classification decisions. Our method first integrates ROI semantics via global self-attention to enhance node representations, enabling each brain region to capture whole-brain context; it then constructs multi-scale hypergraphs to explicitly model functional subnetworks and multi-ROI interactions, overcoming the locality limitations of traditional GNNs and capturing high-order statistical dependencies; finally, a decision-level semantic alignment mechanism selectively injects patient-specific textual embeddings into graph representations, allowing semantics to directly influence predictions without disrupting the original structure. Experiments on public brain network datasets demonstrate state-of-the-art performance, with improved stability and interpretability, particularly in small-sample scenarios.
\end{abstract}


\section{Introduction}
%Functional magnetic resonance imaging (fMRI) provides a non-invasive means to characterize functional brain activity and connectivity, and has become a central tool for brain disorder analysis\cite{feng2022review,smith2004overview,logothetis2008we}. By modeling functional connectivity as brain networks, fMRI-based methods enable systematic investigation of inter-regional interactions, providing critical diagnostic insights for neurodevelopmental disorders such as Autism Spectrum Disorder (ASD) and Attention-Deficit/Hyperactivity Disorder (ADHD). However, complex high-order dependencies, inter-subject variability, and the untapped potential of clinical textual semantics make learning discriminative features extremely challenging.
Functional magnetic resonance imaging (fMRI) is a pivotal non-invasive tool for brain network analysis \cite{jin2025hypergraph, xue2026role, liu2025medsam3}. By modeling functional connectivity as brain networks, it enables systematic investigation of inter-regional interactions, offering critical diagnostic insights for Autism Spectrum Disorder (ASD) and Attention-Deficit/Hyperactivity Disorder (ADHD). However, complex high-order dependencies, inter-subject variability, and underutilized clinical semantics hinder the extraction of discriminative features.
%功能磁共振成像 (fMRI) 为大脑功能活动及其连接的非侵入性刻画提供了手段，已成为脑疾病分析的核心工具。通过将功能连接建模为大脑网络，基于 fMRI 的方法能够对区域间的相互作用进行系统性研究。然而，由于复杂的组织高阶依赖性和显著的个体差异，从中学习判别性特征仍极具挑战。关键在于，现有方法往往忽视了文本信息所提供的关键语义背景，这在标注数据稀缺的情况下，进一步加剧了实现稳健分析的难度。

Since brain networks can be naturally modeled as graph structures, GNNs have emerged as a paradigm for brain modeling. Representative methods like BrainGNN \cite{li2021braingnn} and DHGFormer \cite{xue2025dhgformer} have advanced the analysis of functional connectivity. To further capture complex interactions, hypergraph-based extensions such as I$^2$HGC \cite{han2024inter} have been introduced to model high-order correlations among brain regions. Most graph frameworks overlook the alignment of structural interactions with clinical semantics, failing to leverage semantic intervention for discriminative learning.

% Evolving from CNN-based BrainNetCNN \cite{kawahara2017brainnetcnn}, graph methods like BrainGNN\cite{li2021braingnn}  and PopuDet\cite{11209849} enabled non-Euclidean modeling. However, PopuDet reduces clinical metadata to mere numerical clusters, failing to link ROI semantics with high-order brain topology. To capture complex relations, hypergraph approaches such as I$^2$HGC \cite{han2024inter} further integrate high-order correlations. 
% Recently, Brain Network Transformer\cite{kan2022brain} and Com-BrainTF\cite{bannadabhavi2023community} leveraged Transformers for global dependencies, while frameworks like BrainNPT\cite{hu2024brainnpt}, BrainGSL\cite{wen2023graph}, and DART introduced pre-training to enhance robustness. 
% Nevertheless, even when textual metadata is considered, most methods neglect the intrinsic relationship between high-order brain interactions and the clinical semantics of ROIs, failing to establish a strong semantic intervention to drive the discriminative learning process.

 In practice, clinicians interpret neuroimaging evidence in conjunction with prior knowledge, such as functional roles of brain regions and disease-related semantics. The emergence of Large Language Models (LLMs) offers a new avenue to encode such high-level knowledge and incorporate it into brain network analysis\cite{panagoulias2024evaluating}. Consequently, synergizing high-order brain network modeling with LLM-derived semantics represents an essential research direction.

%There are two main paradigms for combining fMRI with LLMs. The first directly converts fMRI signals into LLM input tokens for text generation or question-answering, as exemplified by fMRI-LM \cite{wei2025fmri} and BP-GPT\cite{chen2024open}. These methods heavily rely on prompt design, with implicit, generative decision processes that can compromise stability and consistency. BP-GPT, for instance, encodes only fMRI time series, ignores high-order dependencies among ROIs, and uses the LLM solely for text generation, without allowing semantic information to directly influence classification or leveraging individual-specific information. The second paradigm uses prompts to guide LLMs in generating disease-related textual information, which is then integrated into the fMRI analysis pipeline, as in BrainPrompt\cite{xu2025brainprompt} and LLM-VEM \cite{ma2024aligned}. While BrainPrompt introduces multi-level prompts to enrich ROI, subject, and disease representations, semantic information remains an auxiliary signal, processed only through local GNNs. Consequently, high-level semantics cannot propagate across the whole brain, multi-ROI functional modules are not fully captured, and semantics do not directly participate in decision-making, limiting robustness in small-sample or specific disease tasks.
\begin{figure}[t]
    \centering
    \includegraphics[width=1.1\columnwidth]{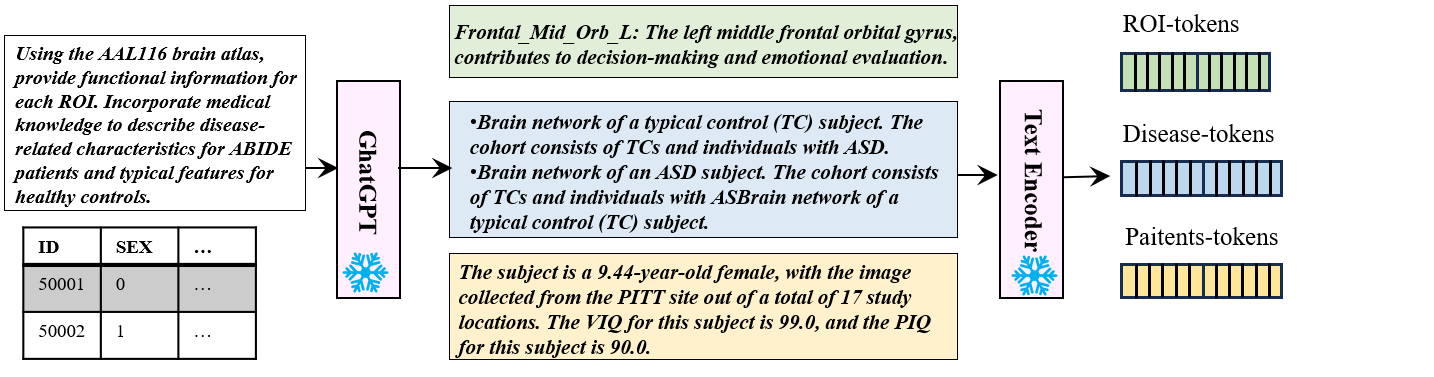}
    \caption{Overview of the LLM-driven pipeline in our proposed Saber framework for generating ROI-level, patient-level, and disease-level semantics, serving as structured semantic priors for subsequent brain network analysis.}
    \label{fig:framework}
\end{figure}

Two main paradigms currently dominate the integration of fMRI with LLMs. The first involves directly mapping fMRI signals into LLM input tokens for generative tasks, as seen in fMRI-LM\cite{wei2025fmri} and BP-GPT\cite{chen2024open}. However, these methods rely heavily on prompt engineering, leading to implicit and often inconsistent decision processes. BP-GPT, for instance, focuses on 1D time series while neglecting high-order interactions among ROIs, treating the LLM merely as a text generator rather than a driver for classification. The second paradigm utilizes LLMs to extract prior knowledge, which is then fused into the imaging pipeline, as in BrainPrompt\cite{xu2025brainprompt} and LLM-VEM\cite{ma2024aligned}. While BrainPrompt introduces multi-level prompts to enrich ROI, subject, and disease representations, semantic information remains an auxiliary signal, processed only through local GNNs. Consequently, high-level semantics cannot propagate across the whole brain, multi-ROI functional modules are not fully captured, and semantics do not directly participate in decision-making, limiting robustness in small-sample or specific disease tasks.

%This limitation raises a core question: can language semantics actively guide the decision-making process of brain network classification, rather than merely enhancing representations or providing supervision? To address this, we propose a semantic-aligned brain network modeling framework that elevates LLM-derived semantics from auxiliary prompts to decision-level reasoning components. Our method captures both local and global high-order interactions through multi-scale brain network modeling and enables direct interaction between graph-level embeddings and individualized language semantics via a decision-level semantic alignment mechanism, allowing semantic cues to actively shape the final prediction.
This limitation raises a fundamental question: can language semantics actively participate in the decision-making process of brain network classification, rather than merely serving as feature augmentation or supervision? 
%To address this question, we propose Saber, a semantic-aligned brain network framework that enables LLM-derived semantics to actively participate in brain network classification. Saber adopts a hierarchical design that progressively integrates semantic priors from ROI-level representations to graph-level abstraction and decision-level prediction, allowing semantic knowledge to guide learning without disrupting the underlying brain topology.
To address this, we propose Saber, a semantic-aligned brain network modeling framework that elevates LLM-derived semantics to decision-level reasoning components. The overall pipeline for generating ROI-level, patient-level, and disease-level semantics from medical texts using LLMs is illustrated in Figure~\ref{fig:framework}. Specifically, our model integrates ROI semantics via global self-attention at the node level. This enables brain regions to perceive both local connectivity and whole-brain semantic priors, yielding enriched node representations for subsequent high-order modeling. Subsequently, multi-scale hypergraphs are constructed to capture multi-ROI co-activations, overcoming GNN locality and pairwise graph constraints. Meanwhile, a gated fusion mechanism balances cross-scale features to extract robust high-order dependencies. Finally, a graph-level semantic alignment mechanism selectively injects patient-specific textual embeddings into the graph representation, enabling semantic information to directly influence the final prediction without disrupting the original structural information. Our hierarchical design drives multi-scale semantic--structural interactions, improving robustness and discrimination in data-scarce tasks.

\begin{figure*}[t]
    \centering
    \includegraphics[width=\textwidth]{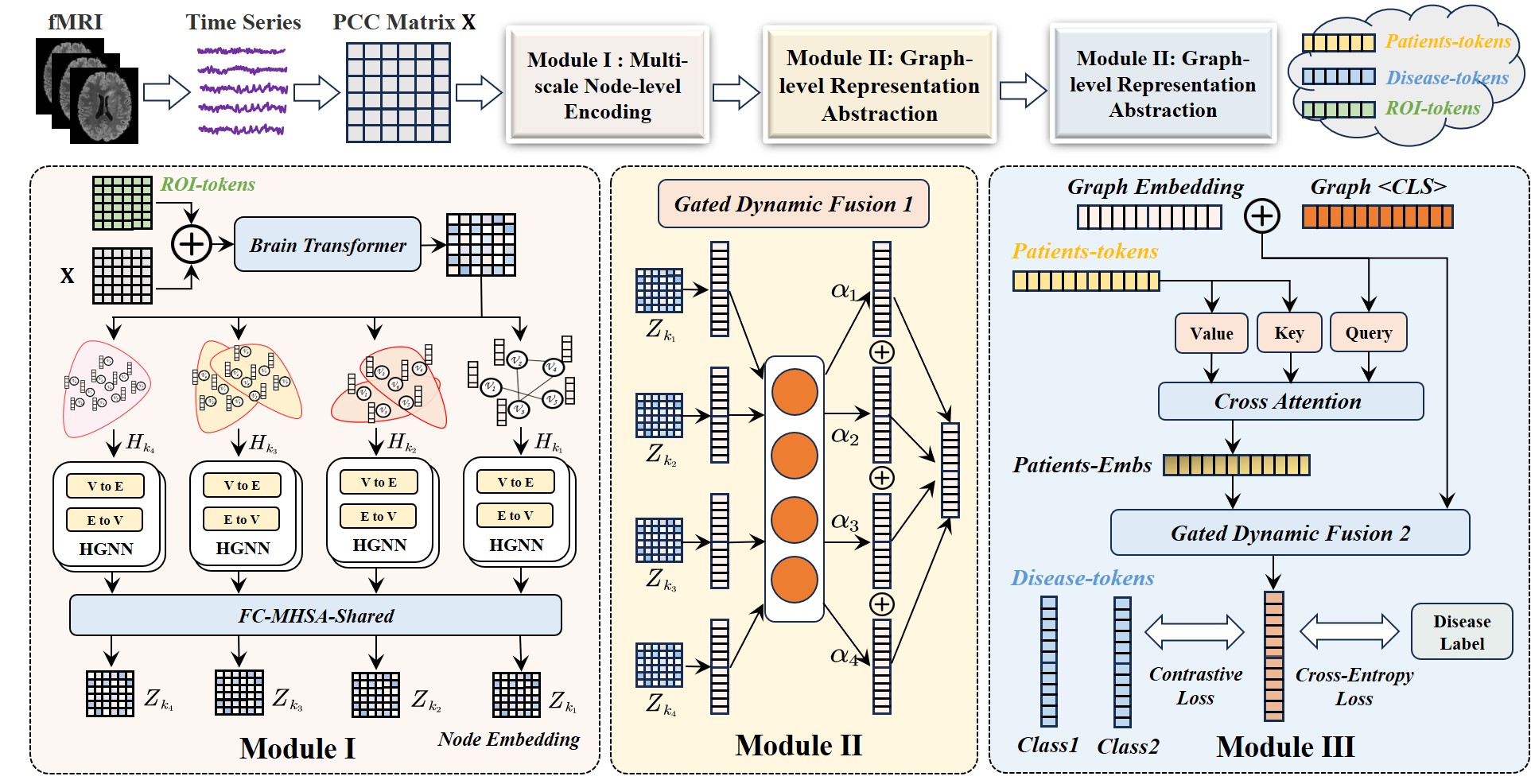}
    \caption{Overview of the proposed LLM-guided brain network framework Saber that progressively injects semantic priors from ROI-level node representations to graph-level abstraction and decision-level prediction.}
    \label{fig:framework2}
\end{figure*}
The main contributions of this work are summarized as:
\begin{itemize}
%\item{We propose a hierarchical framework that synergizes multi-scale hypergraph modeling with semantic alignment, facilitating deep interactions between brain topology and patient-specific semantics at both node and graph levels.}

%\item{We shift the role of LLM-derived clinical knowledge from passive auxiliary prompts to active, decision-level reasoning components, overcoming the limitations of current fMRI-LLM integration.}

\item{We introduce a decision-level semantic alignment mechanism that enables LLM-derived patient semantics to directly guide brain network classification.}

\item{We propose a multi-scale hypergraph-based node-to-graph modeling strategy to capture high-order and multi-ROI interactions beyond pairwise connectivity.}

\item{Extensive evaluations on ABIDE and ADHD-200 show that Saber consistently outperforms existing methods.}
\end{itemize}

\section{Method}
The overall framework of the Saber is illustrated in Figure~\ref{fig:framework2}. In the following, we describe its three stages, progressing from node-level to graph-level representations.
\subsection{Module I: Multi-scale Node-level Brain Network Encoding}
\noindent\textbf{Node Feature Initialization.} The brain is parcellated into $N$ Regions of Interest (ROIs) utilizing the AAL atlas \cite{desikan2006automated}. We compute the Pearson correlation between the BOLD signals to derive the initial vertex feature matrix $\mathbf{X} \in \mathbb{R}^{N \times N}$.

\noindent\textbf{ROI Semantic Embedding Injection.} 
To incorporate clinical semantic, we generate detailed textual descriptions for each ROI using LLM, focusing on anatomical locations and functional roles. These descriptions are encoded by a frozen pre-trained text encoder to yield semantic embeddings $\mathbf{E}_{\text{ROI}} \in \mathbb{R}^{N \times d_{\text{LLM}}}$. Subsequently, $\mathbf{E}_{\text{ROI}}$ is projected into the node feature space via a learnable linear projection $\mathbf{W}_p$ and fused with the original connectivity-based features $\mathbf{X}$. To facilitate global semantic propagation across regions, we employ the encoder of the Brain Network Transformer (BNT)~\cite{kan2022brain}:
% We construct ROI-specific textual descriptions using ChatGPT based on anatomical locations and functional roles, and encode them with a frozen pre-trained text encoder to obtain semantic embeddings $\mathbf{E}_{\text{ROI}} \in \mathbb{R}^{N \times d_{LLM}}$. The embeddings are projected into the node feature space via a learnable projection $\mathbf{W}_p$ and fused with connectivity-based features $\mathbf{X}$. The brain net transformer \cite{kan2022brain} encoder is employed to enable global semantic propagation across ROIs:
\begin{equation}
\mathbf{X}_0 = \text{BNT}(\mathbf{X} + \mathbf{E}_{\text{ROI}}\mathbf{W}_p),
\end{equation}
% where $\mathbf{X}_0 \in \mathbb{R}^{N \times N}$ denotes the semantic-augmented vertex representations.  This global self-attention mechanism allows each ROI to jointly capture subject-specific connectivity and high-level functional priors, forming the basis for subsequent multi-scale hypergraph modeling.
This global self-attention mechanism enables jointly capturing brain connectivity patterns and high-level functional priors, forming the basis for multi-scale hypergraph modeling.

\noindent\textbf{Multi-scale Hypergraph Modeling.}  
Based on the semantic-augmented node features $\mathbf{X}_0 $, we construct multi-scale hypergraphs to capture high-order interactions among ROIs. At each scale $k$, a hypergraph $\mathcal{H}^{(k)} = (\mathcal{V}, \mathcal{E}^{(k)})$ is defined with incidence matrix $\mathbf{H}^{(k)} \in \mathbb{R}^{N \times M_k}$:
\begin{equation}
\mathbf{H}^{(k)}_{i,e} =
\begin{cases}
1, & \text{if ROI } i \text{ belongs to hyperedge } e,\\
0, & \text{otherwise},
\end{cases}
\end{equation}
where $M_k$ represents the number of hyperedges at scale $k$. Each hyperedge links a group of ROIs sharing similar semantic-enriched functional characteristics, enabling modeling beyond pairwise connectivity. The ensemble of hypergraphs at multiple scales $\{k_1,\dots,k_K\}$ jointly encapsulates both local and global interaction patterns.

For each scale $k$, vertex features are propagated through a two-layer Hypergraph Neural Network (HGNN).
\begin{equation}
\mathbf{X}^{(k,l+1)} = \sigma\!\left(
\mathbf{D}_v^{-\frac{1}{2}} \mathbf{H}^{(k)} \mathbf{D}_e^{-1} (\mathbf{H}^{(k)})^\top \mathbf{D}_v^{-\frac{1}{2}}
\mathbf{X}^{(k,l)} \mathbf{W}^{(k,l)}
\right),
\end{equation}
where $\mathbf{X}^{(k,0)}=\mathbf{X}_0$, $\mathbf{D}_v$ and $\mathbf{D}_e$ are degree matrices, and $\sigma(\cdot)$ is an activation function. The resulting vertex features at each scale are further refined by a shared Multi-Head Self-Attention (MHSA) module. This module facilitates the transition of information from fine-grained local structures to coarser global patterns while maintaining consistency across scales:
\begin{equation}
\mathbf{Z}^{(k)} = \text{MHSA}\!\left(\mathbf{X}^{(k,2)}\right).
\end{equation}

By sharing the attention parameters across scales, the model enforces a unified interaction pattern for integrating multi-scale representations, preventing scale-specific overfitting and enabling coherent aggregation in the subsequent graph-level abstraction module. The resulting multi-scale node representations $\{\mathbf{Z}^{(k)}\}_{k=1}^{K}$ are then forwarded to Module II.

\subsection{Module II: Graph-level Representation Abstraction}

Given the multi-scale node-level representations $\{\mathbf{Z}^{(k)}\}_{k=1}^{K}$ yielded by Module I, this module synthesizes them into a unified graph-level representation. This process captures subject-specific brain network characteristics while preserving complementary information from varying interaction scales.

\noindent\textbf{In-scale Node-to-Graph Pooling.}
For each hypergraph scale $k$, node-level features $Z^{(k)}$ are aggregated into a scale-specific graph representation via global average pooling:
\begin{equation}
\mathbf{g}^{(k)} = \frac{1}{N} \sum_{i=1}^{N} \mathbf{Z}^{(k)}_i,
\end{equation}
where $\mathbf{g}^{(k)}$ summarizes the overall functional organization of the brain network under scale $k$, providing a compact global descriptor while preserving scale-specific characteristics.

\noindent\textbf{Adaptive Multi-scale Fusion.}
% As the relevance of different neighborhood scales may vary across subjects, we employ a learnable gating mechanism to adaptively fuse scale-specific graph representations.
Recognizing that the contribution of different interaction scales varies across subjects, we employ a learnable gating mechanism to adaptively fuse scale-specific representations.
Specifically, the pooled features from all scales are concatenated and passed to a gating network:
\begin{equation}
\boldsymbol{\alpha} = \text{softmax}\left( f_{\text{gate}}\left([\mathbf{g}^{(1)} \| \mathbf{g}^{(2)} \| \cdots \| \mathbf{g}^{(K)}]\right) \right),
\end{equation}
% where $\boldsymbol{\alpha} \in \mathbb{R}^{K}$ denotes subject-specific importance weights for different scales. The final graph-level representation is then obtained as a weighted combination:
where $\boldsymbol{\alpha} \in \mathbb{R}^{K}$ denotes the subject-specific importance weights for different scales, and $\|$ represents the concatenation operation. The final graph-level representation is derived as a weighted combination:
\begin{equation}
\mathbf{g} = \sum_{k=1}^{K} \alpha_k \mathbf{g}^{(k)}.
\end{equation}

The resulting representation $\mathbf{g}$ provides a compact yet discriminative summary of the brain network and is forwarded to Module III for decision-level semantic enhancement.

\subsection{Module III: Semantic Alignment and Decision-level Enhancement}
Module III advances the integration of LLM-derived semantics from auxiliary feature augmentation to explicit decision-level reasoning. By aligning patient-specific textual information and label semantics with the graph-level representation $\mathbf{g}$, this module allows semantic knowledge to guide predictions while preserving the structural integrity of the brain network.

\noindent\textbf{Patient-specific Semantic Injection.}  
 % Unlike conventional CLS tokens, it is combined with the graph-level embedding $\mathbf{g}$ before attending LLM text, forming the query
To enable controlled decision-level reasoning, we introduce a dedicated \textit{Graph-CLS} token $\mathbf{q}_{\text{CLS}} \in \mathbb{R}^{d}$, which is decoupled from the structural pooling process. Unlike conventional CLS tokens, $\mathbf{q}_{\text{CLS}}$ is fused with the graph-level embedding $\mathbf{g}$ prior to attending to the LLM text, forming a conditioned query:
\begin{equation}
\mathbf{q} = \mathbf{q}_{\text{CLS}} + \mathbf{g}.
\end{equation}

This design ensures that semantic retrieval is explicitly conditioned on the graph representation. It enables the model to extract patient-specific semantic cues that are contextually relevant to the brain network, thereby filtering out irrelevant or noisy information, which is a key innovation of this module.

Patient metadata, including demographics and clinical descriptions, are encoded via a pre-trained LLM into a sequence of embeddings $\mathbf{E}_{\text{Patient}} \in \mathbb{R}^{N_{\text{text}} \times d_{\text{LLM}}}$. The combined token $\mathbf{q}$ then attends to these embeddings through cross-attention:
\begin{equation}
\mathbf{h}_{\text{fused}} = \text{CrossAttn}(\mathbf{q}, \mathbf{E}_{\text{Patient}}, \mathbf{E}_{\text{Patient}}),
\end{equation}
This mechanism allows the model to selectively attend to the most informative patient-specific semantic cues for the prediction task, preventing unfiltered semantic noise from corrupting the graph representation.

\noindent\textbf{Gated Residual Semantic Injection.}  
To regulate semantic contribution, we employ a gated residual fusion mechanism:
\begin{equation}
\mathbf{g}' = \mathbf{g} + \sigma(\mathbf{W}_g \mathbf{h}_{\text{fused}}) \odot \mathbf{h}_{\text{fused}},
\end{equation}
where $\sigma(\cdot)$ is sigmoid and $\odot$ is element-wise multiplication. This adaptive gating dynamically modulates the strength of semantic injection per subject, ensuring robustness against noisy or irrelevant textual information.

\noindent\textbf{Decision-level Supervision.}  
The semantically enriched representation $\mathbf{g}'$ is optimized using a dual-supervision strategy. First, standard cross-entropy loss enforces accuracy:
\begin{equation}
\mathcal{L}_{\text{CE}} = - \log \hat{p}_y, \quad \hat{\mathbf{p}} = \text{Softmax}(\text{MLP}(\mathbf{g}')).
\end{equation}

Second, a contrastive loss aligns $\mathbf{g}'$ with LLM-derived disease embeddings $\mathbf{E}_{\text{Disease}} = \{\mathbf{e}_1, \dots, \mathbf{e}_C\}$, encouraging proximity to the corresponding class embedding $\mathbf{e}_y$ while pushing away from others:
\begin{equation}
\mathcal{L}_{\text{CL}} = - \log \frac{\exp(\text{sim}(\mathbf{g}', \mathbf{e}_y)/\tau)}{\sum_{c=1}^{C} \exp(\text{sim}(\mathbf{g}', \mathbf{e}_c)/\tau)}.
\end{equation}

The total training objective is
\begin{equation}
\mathcal{L}_{\text{total}} = \mathcal{L}_{\text{CE}} + \lambda \mathcal{L}_{\text{CL}}.
\end{equation}

\section{Experiments}

\begin{table*}[ht]
\centering
\caption{Comparison on ABIDE and ADHD datasets. Best results are \textbf{bolded}. All values are reported as mean $\pm$ standard deviation (\%). SABER$_{\text{lite}}$ denotes the variant without LLM-based semantic modeling, while SABER represents the full model.}
\label{tab:sota_comparison}
\renewcommand{\arraystretch}{1.2}
\setlength{\tabcolsep}{4pt}
\resizebox{\textwidth}{!}{
\begin{tabular}{lcccccccc}
\toprule
\multirow{2.5}{*}{\textbf{Method}} & \multicolumn{5}{c}{\textbf{ABIDE Dataset}} & & \multicolumn{2}{c}{\textbf{ADHD Dataset}} \\
\cmidrule(lr){2-6} \cmidrule(lr){8-9}
& \textbf{Accuracy} & \textbf{AUC} & \textbf{Recall} & \textbf{Specificity} & \textbf{F1 Score} & & \textbf{Accuracy} & \textbf{AUC} \\
\midrule
BrainNetCNN \cite{kawahara2017brainnetcnn}     
& $65.81 \pm 1.91$ & $70.22 \pm 2.66$ & $69.03 \pm 3.96$ & $65.97 \pm 2.51$ & $67.33 \pm 1.25$ & 
& $61.42 \pm 1.87$ & $62.03 \pm 2.14$ \\

BrainGNN \cite{li2021braingnn}        
& $64.43 \pm 1.66$ & $68.14 \pm 1.69$ & $65.31 \pm 5.44$ & $63.81 \pm 3.79$ & $64.41 \pm 3.69$ & 
& $60.95 \pm 2.02$ & $61.88 \pm 2.36$ \\

BNT \cite{kan2022brain}             
& $66.13 \pm 2.36$ & $71.34 \pm 3.95$ & $71.87 \pm 6.96$ & $67.41 \pm 3.71$ & $69.33 \pm 3.31$ & 
& $62.87 \pm 1.74$ & $63.45 \pm 2.01$ \\

Com-BrainTF \cite{bannadabhavi2023community}     
& $64.41 \pm 2.89$ & $70.77 \pm 2.85$ & $69.03 \pm 6.12$ & $66.35 \pm 3.22$ & $67.44 \pm 3.16$ & 
& $62.11 \pm 2.28$ & $62.74 \pm 2.47$ \\

HGNN \cite{feng2019hypergraph}            
& $64.52 \pm 2.10$ & $70.41 \pm 2.37$ & $66.72 \pm 5.81$ & $65.15 \pm 3.90$ & $65.67 \pm 2.59$ & 
& $61.76 \pm 1.93$ & $62.20 \pm 2.10$ \\

BrainNPT \cite{hu2024brainnpt}        
& $64.32 \pm 3.30$ & $69.47 \pm 4.33$ & $66.39 \pm 4.11$ & $65.38 \pm 4.97$ & $65.57 \pm 0.81$ & 
& $61.58 \pm 2.41$ & $62.09 \pm 2.68$ \\

BrainGSL \cite{wen2023graph}        
& $66.31 \pm 4.38$ & $67.35 \pm 5.55$ & $66.82 \pm 5.58$ & $67.57 \pm 5.57$ & $66.93 \pm 3.87$ & 
& $63.24 \pm 2.35$ & $63.88 \pm 2.79$ \\

MADEforASD \cite{liu2024made}           
& $65.21 \pm 2.28$ & $69.49 \pm 1.28$ & $69.69 \pm 4.77$ & $65.09 \pm 3.80$ & $67.13 \pm 2.22$ & 
& $62.63 \pm 1.96$ & $63.14 \pm 2.22$ \\

BrainPrompt \cite{xu2025brainprompt}    
& $66.35 \pm 7.14$ & $66.45 \pm 6.55$ & $71.08 \pm 13.42$ & $68.38 \pm 8.54$ & $68.72 \pm 8.11$ & 
& $64.92 \pm 1.88$ & $65.37 \pm 2.04$ \\

\midrule
SABER$_{\text{lite}}$            
& $69.82 \pm 1.34$ & \textbf{72.23 $\pm$ 1.15} & $71.97 \pm 4.83$  & $68.88 \pm 4.14$ & $70.02 \pm 3.06$ & 
& $65.12 \pm 2.42$ & $65.68 \pm 2.77$ \\

SABER     
& \textbf{71.58 $\pm$ 4.08} & $70.40 \pm 3.02$ & \textbf{81.87 $\pm$ 3.02} & \textbf{69.68 $\pm$ 6.25} & \textbf{75.06 $\pm$ 5.82} & 
& \textbf{66.21 $\pm$ 2.31} & \textbf{66.87 $\pm$ 2.59} \\
\bottomrule
\end{tabular}
}
\end{table*}

\subsection{Datasets}

\subsubsection{ABIDE Dataset} We evaluated our proposed method on the Autism Brain Imaging Data Exchange (ABIDE I) dataset\cite{di2014autism}. ABIDE I consists of rs-fMRI data collected from 16 international sites. To ensure data quality, we selected 871 quality-checked subjects, comprising 403 patients with Autism Spectrum Disorder (ASD) and 468 Normal Controls (NC).

\subsubsection{ADHD-200 Dataset} We further validated our model using the ADHD-200 dataset, a large-scale public database aggregated from eight imaging centers. This dataset includes 362 patients diagnosed with Attention Deficit Hyperactivity Disorder (ADHD) and 585 NCs. 
Both datasets were preprocessed using the DPARSF pipeline \cite{yan2010dparsf}. 
% Regional time series were extracted based on the AAL-116 atlas \cite{desikan2006automated} by averaging voxel-level fMRI signals within each ROI. Subsequently, functional connectivity matrices ($116 \times 116$) were constructed via pairwise Pearson correlation coefficients, providing the input for the following graph and hypergraph modeling.

To ensure robust training and evaluation, we employed a stratified sampling strategy to partition the dataset into training, validation, and test sets with an 8:1:1 ratio. This stratification preserves the class distribution across splits, ensuring a balanced representation of patients and normal controls.
\subsection{Evaluation Setups}
\subsubsection{Implementation Details.}
Model training was performed on a single NVIDIA RTX 3090 GPU with 24GB of memory. The batch size was set to 16, and the model was trained for 100 epochs using the AdamW optimizer. We adopted an initial learning rate of $1 \times 10^{-4}$ and applied weight decay of $1 \times 10^{-4}$ to prevent overfitting and enhance generalization. A cosine annealing learning rate scheduler was used during training to improve convergence. To further validate the model's reliability and generalizability, we also conducted ten-fold cross-validation.

For text generation and encoding, we use  ChatGPT-5 and Llama-encoder-1.0B model from the LLM2Vec framework as a frozen encoder. ROI, patients, and disease text are encoded in a $d_{llm}$ dimensional embedding, which is then linearly projected onto a 116-dimensional vector matching the node features. We evaluated hypergraphs across multiple scales ($k \in \{1, 5, 10, 15\}$) to capture multi-resolution structural representations of brain dynamics. These localized and global features are fused within a unified framework. Based on empirical performance, we set the number of attention heads to 8. 

%In order to explore the effect of structural granularity in brain networks, we constructed hypergraphs at multiple neighborhood resolutions. Specifically, we evaluated three different $k$-nearest neighbor configurations: $k=1, 5, 10$, and $15$, representing fine, medium, and coarse-grained functional connectivity contexts, respectively. These settings allow our model to learn structural representations at varying scales, which are later fused in a unified framework to capture both localized and global brain dynamics relevant to ASD. The number of attention heads in the multi-stage Transformer is set to 8. The above parameters were selected to achieve optimal performance through empirical experimentation. Specifically, the selection process and experimental results for key hyperparameters, such as the number of attention heads in the Transformer, the number of training epochs, and the batch size will be elaborated in the following sections.

\subsubsection{Baseline Methods}
We evaluated the proposed model against several state-of-the-art baselines, including BrainNetCNN\cite{kawahara2017brainnetcnn}, HGNN\cite{feng2019hypergraph}, BrainNetTransformer\cite{kan2022brain}, BrainGNN\cite{li2021braingnn} , Com-BrainTF\cite{bannadabhavi2023community} , MADEforASD\cite{liu2024made} , BrainGSL\cite{wen2023graph} and BrainPrompt\cite{xu2025brainprompt}. These methods represent a diverse range of brain network modeling strategies. 

\subsubsection{Evaluation Metrics}Model performance was assessed using Accuracy, AUC, Recall, Specificity, and F1-score to provide a comprehensive multi-perspective evaluation.

\subsection{Classification Performance Results and Analysis}

As illustrated in Table~\ref{tab:sota_comparison}, the proposed Saber model achieves state-of-the-art performance on both ABIDE (attaining 71.58 accuracy) and ADHD-200 datasets. These results validate our core hypothesis: elevating clinical semantics from auxiliary features to decision-level components significantly bolsters the robustness of brain network analysis.

\subsubsection{Structural Superiority} Compared to Transformer-based architectures such as BrainNetTransformer and Com-BrainTF, Saber's advantage stems from its Multi-scale Hypergraph Modeling (Module I). While standard Transformers struggle to capture the non-Euclidean topology of brain networks, our approach extracts high-order interactions that align more closely with the brain's intrinsic functional organization. Notably, even the Saber$_{\text{lite}}$ (the version without LLM enhancement) consistently outperforms most baselines, including self-supervised frameworks like BrainNPT, BrainGSL. 
\subsubsection{Synergy of Semantic-Decision Alignment} Saber demonstrates a substantial leap in classification accuracy and stability over BrainPrompt. While BrainPrompt primarily treats semantics as local GNN inputs, Saber's Graph-CLS Token acts as a decision-level query, selectively extracting critical evidence from patient-specific texts via cross-attention. This design directly addresses the fundamental limitation where semantics fail to participate in the final decision-making process. Through Gated Residual Injection, clinical text evolves from a passively attached label into an active constraint that shapes the classification boundary. The consistent performance across both datasets highlights Saber's exceptional adaptability; by synergizing high-order imaging features with textual priors, the model effectively emulates the "imaging + clinical records" dual-verification logic employed by medical experts.

\begin{table}[t]
\centering
\caption{Ablation study on different architectural components.}
\label{tab:ablation_two_datasets}
\renewcommand{\arraystretch}{1.1}
\setlength{\tabcolsep}{6pt}
\begin{tabular}{lcccc}
\toprule
\multirow{2}{*}{\textbf{Configuration}} 
& \multicolumn{2}{c}{\textbf{ABIDE}} 
& \multicolumn{2}{c}{\textbf{ADHD-200}} \\
\cmidrule(lr){2-3} \cmidrule(lr){4-5}
& \textbf{Acc} & \textbf{AUC} & \textbf{Acc} & \textbf{AUC} \\
\midrule
w/o TransEnc          & 61.84 & 66.75 & 60.92 & 63.41 \\
single-scale (k=10)  & 61.36 & 64.84 & 60.15 & 62.87 \\
double-scale (k=5,10)     & 64.22 & 69.25 & 63.01 & 66.38 \\
MHSA-noshared             & 63.92 & 68.97 & 62.48 & 65.92 \\
w/o Dynamic Fusion        & 64.02 & 68.82 & 62.77 & 65.41 \\
w/o Dynamic Fusion \& CLS & 64.72 & 69.74 & 63.35 & 66.02 \\
w/o CLS                   & 63.43 & 69.52 & 62.11 & 65.76 \\
w/o ROI Semantic Injection & 70.28 & 71.43 & 65.17 & 65.59 \\
w/o Patient Semantic Injection & 70.11 & 69.89 & 64.98 & 65.09\\
w/o Disease Contrastive& 68.47 & 68.94 & 64.12 & 64.79\\
\midrule
SABER$_{\text{lite}}$    &69.82 & \textbf{72.23} & 65.12 & 65.68 \\
SABER &  \textbf{71.58} &70.40 &  \textbf{66.21 }& \textbf{66.97} \\
\bottomrule
\end{tabular}
\end{table}

\subsection{Ablation Study}

Table \ref{tab:ablation_two_datasets} shows ablation results on ABIDE and ADHD-200.

\subsubsection{Multi-scale structure $\&$ shared attention} Removing the Transformer or using a single scale ($k=10$) degrades performance, indicating that fixed-scale modeling fails to capture high-order brain interactions. We therefore adopt a multi-scale setting with $k \in \{1, 5, 10, 15\}$ to hierarchically model structures from local ROIs to large-scale inter-regional coordination. In contrast, disabling shared attention (MHSA-noshared) disrupts cross-scale alignment and reduces performance consistency.

%Removing the Transformer or using a single scale ($k=10$) drops performance, highlighting the need for high-order interactions and cross-scale feature integration. Dual scales ($k=5,10$) improve Acc and AUC, while disabling shared attention (MHSA-noshared) reduces consistency.
\subsection{Impact of Key Hyperparameters on Model Performance}
\begin{figure}[htbp]
  \centering
  \begin{minipage}{0.48\columnwidth}
    \centering
    \includegraphics[width=\textwidth]{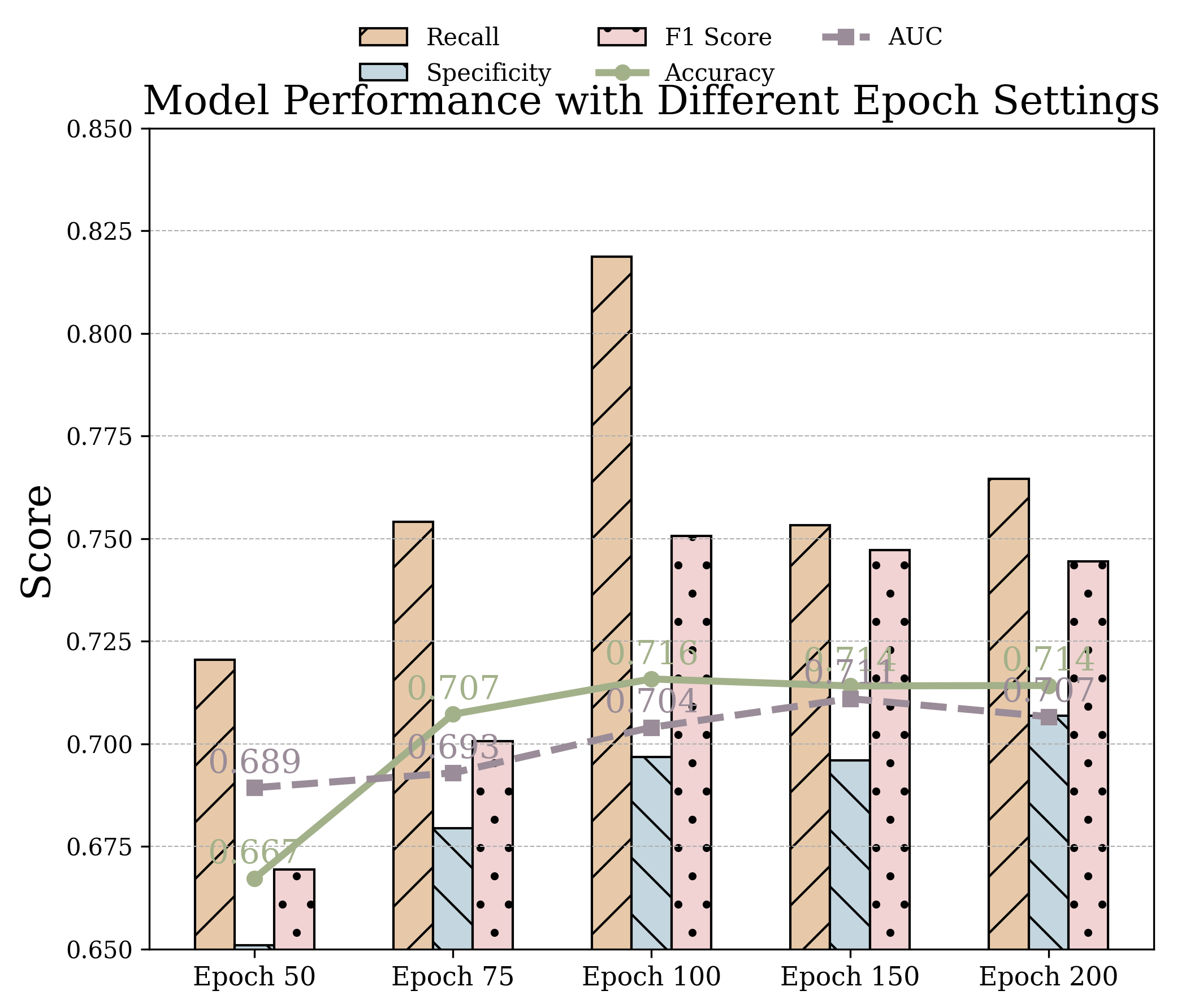}
    \caption{Perf. vs. Epochs.}
    \label{fig:epoch_performance}
  \end{minipage}
  \hfill 
  \begin{minipage}{0.48\columnwidth}
    \centering
    \includegraphics[width=\textwidth]{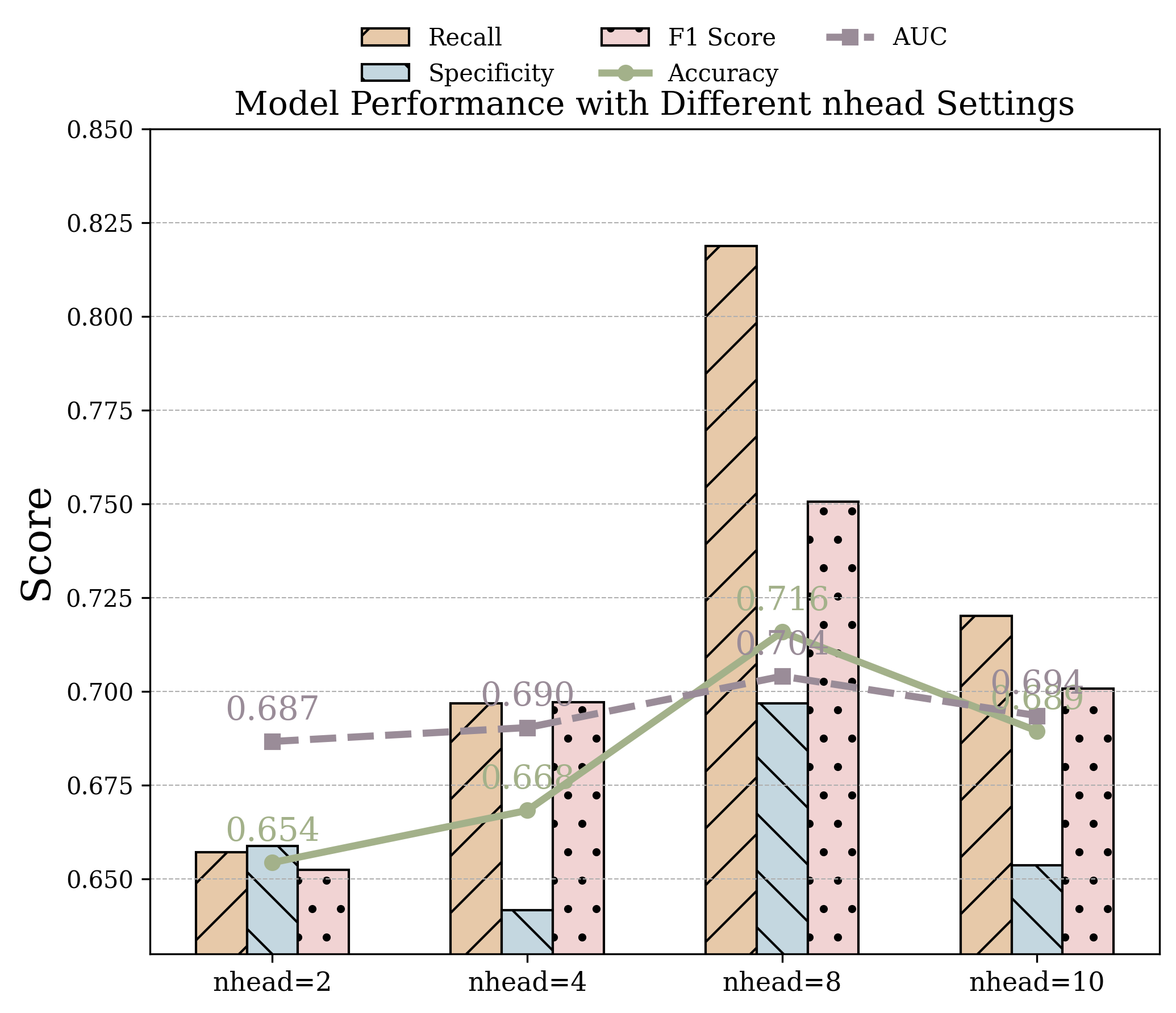}
    \caption{Effect of Heads.}
    \label{fig:atthead}
  \end{minipage}
\end{figure}
\subsubsection{Dynamic Fusion $\&$ Graph-CLS} Excluding dynamic fusion weakens performance, indicating adaptive multi-scale integration is crucial. Removing Graph-CLS or both CLS and dynamic fusion further lowers results, showing decision-level queries aid semantic selection without harming structural representation.

\subsubsection{Hierarchical semantic alignment} Removing ROI- or patient-level semantic injection reduces performance, and omitting disease-level semantic contrast significantly harms discriminative power, confirming the complementary role of multi-level semantic priors.

Full SABER consistently outperforms SABER$_{\text{lite}}$, demonstrating that decision-level semantic enhancement stabilizes performance and improves generalization.

% \begin{figure}[htbp]
%   \centering
%   \includegraphics[width=0.45\textwidth]{nhead_performance_plot.png}
%   \caption{Effect of varying attention head numbers on model performance.}
%   \label{fig:atthead}
% \end{figure}

% To maintain consistency, the number of attention heads was fixed across Transformer layers. Ablation studies (Figure~\ref{fig:atthead}) show performance improves from 2 to 8 heads but slightly drops at 10. This suggests that multiple heads enhance feature diversity and interaction modeling, while too many heads can fragment the feature space. A moderate head count thus balances representation capacity and stability.

% We further evaluated model performance under different training durations: 50, 75, 100, 150, and 200 epochs. As illustrated in Figure~\ref{fig:epoch_performance}, performance metrics generally improved with increasing epochs, reaching their peak---accuracy, AUC, and F1 score---at 100 epochs. Beyond this point, a slight decline was observed, likely due to overfitting and reduced generalization. These results highlight the importance of selecting an appropriate number of training epochs; based on our experiments, 100 epochs was identified as the optimal setting.
We also evaluated different training durations (50--200 epochs). As shown in Figure~\ref{fig:epoch_performance}, performance peaks at 100 epochs, with longer training causing slight declines, likely due to overfitting. This identifies 100 epochs as the optimal setting.

The number of attention heads was fixed across Transformer layers for consistency. Ablation studies (Figure~\ref{fig:atthead}) show performance improves from 2 to 8 heads but slightly drops at 10, indicating that a moderate head count balances feature diversity and stability.

\renewcommand{\arraystretch}{1.2}
\begin{table}[t]
\centering
\caption{Performance comparison under different batch sizes (\%)}
\label{tab:batchsize_performance}
\begin{tabular}{|c|c|c|c|c|c|}
\hline
\textbf{Batch Size} & \textbf{Accuracy} & \textbf{AUC} & \textbf{Recall} & \textbf{Specificity} & \textbf{F1} \\
\hline
4  & 62.38 & 67.10 & 66.35 & 62.73 & 64.49 \\
8  & 65.89 & 67.25 & 73.23 & 65.50 & 66.57 \\
16 & 71.58 & 70.04 & 81.87 & 69.68 & 75.06 \\
32 & 64.85 & 66.04 & 70.19 & 64.60 & 67.28 \\
64 & 63.37 & 66.33 & 72.12 & 62.50 & 66.96 \\
\hline
\end{tabular}
\end{table}

We evaluated the effect of different batch sizes on model performance (Table~\ref{tab:batchsize_performance}). Smaller batches improved sensitivity, while larger batches enhanced training stability, each with trade-offs in efficiency or generalization. Balancing performance and training time, we chose a batch size of 16 for optimal efficiency and accuracy.

\begin{figure}[t]
    \centering
    \includegraphics[width=0.8\columnwidth]{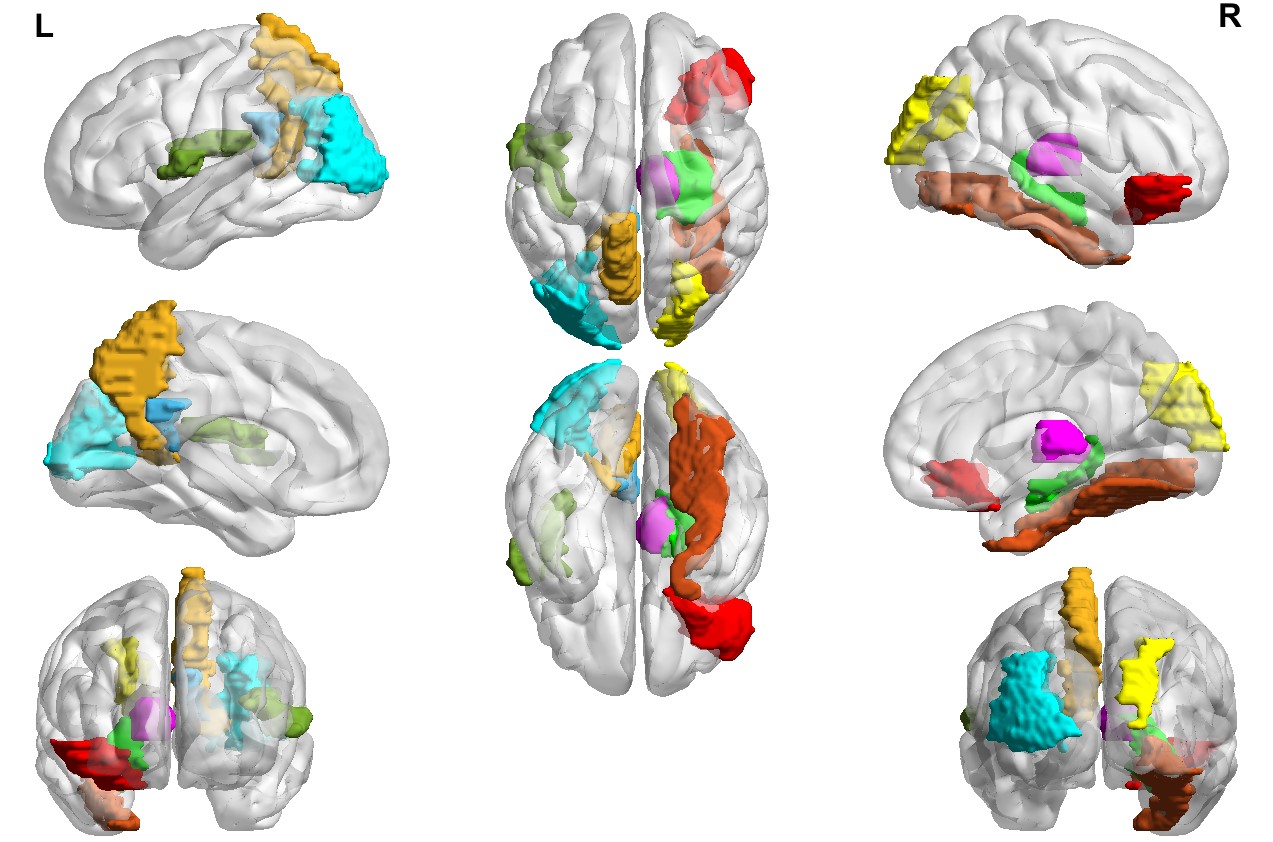}
    \caption{The visualization on discriminative ROIs for ASD Diagnosis.}
    \label{fig:framework}
\end{figure}

\subsection{Interpretability Analysis of Discriminative ROIs}
To improve model interpretability and explore its neuroscientific relevance, we analyzed the most discriminative functional connections and ROIs for ASD classification. Using individualized intra-hypergraph representations, we constructed subject-specific FC matrices. Key connections were identified via a combination of two-sample t-tests and Lasso regression, ensuring both statistical significance and sparsity. The analysis highlighted that the most discriminative connections predominantly involve the fusiform gyrus, hippocampus, inferior frontal gyrus (orbital part), middle occipital gyrus, and posterior cingulate cortex. These regions, frequently implicated in significant connections, may serve as potential neurobiological biomarkers for ASD, consistent with prior findings in the literature~\cite{han2025hypergraph} (see Fig.~\ref{fig:framework}).

\section{Conclusion}

In this paper, we proposed Saber, a semantic-aligned brain network modeling framework that integrates LLM-derived textual knowledge with multi-scale fMRI-based hypergraph representations. By progressively injecting semantic priors from ROI-level node features to graph-level abstraction and decision-level prediction, Saber enables patient-specific semantics to actively guide classification. Multi-scale hypergraph modeling captures high-order, cross-ROI interactions beyond pairwise connectivity, while adaptive fusion and gated semantic injection ensure robust integration of structural and semantic information. Extensive experiments on ABIDE and ADHD-200 demonstrate that Saber consistently outperforms existing methods, offering improved accuracy, robustness, and interpretability. Furthermore, analysis of discriminative functional connections and ROIs reveals neurobiologically meaningful patterns, highlighting the potential of combining high-order brain network modeling with LLM-driven semantic knowledge for advancing neurodevelopmental disorder analysis.

% \section*{Acknowledgment}

% The preferred spelling of the word ``acknowledgment'' in America is without 
% an ``e'' after the ``g''. Avoid the stilted expression ``one of us (R. B. 
% G.) thanks $\ldots$''. Instead, try ``R. B. G. thanks$\ldots$''. Put sponsor 
% acknowledgments in the unnumbered footnote on the first page.
\bibliographystyle{IEEEtran}
\bibliography{paper}

\end{document}